\documentclass[journal]{IEEEtran}

\usepackage{ulem}
\usepackage{times}
\usepackage{epsfig}
\usepackage{amsmath}
\usepackage{amssymb}
\usepackage{amsmath}

\usepackage[pagebackref=true,breaklinks=true,letterpaper=true,colorlinks,bookmarks=false]{hyperref}

\usepackage{enumerate}
\usepackage{multirow}  
\usepackage{stfloats}
\usepackage{booktabs}
\usepackage{diagbox}

\usepackage{graphics}
\usepackage{subfigure}
\graphicspath{{img/}}

\usepackage{algorithm}
\usepackage{algorithmic}

\begin{document}

\title{Few-shot Object Detection on Remote Sensing Images}

\author{Jingyu Deng{*}, Xiang Li{*}, and Yi Fang
\thanks{Xiang Li and Yi Fang are with Multimedia and Visual Computing Lab, NYU Tandon and Abu Dhabi, Department of Electrical and Computer Engineering, NYU Tandon, USA, and Department of Electrical and Computer Engineering, NYU Abu Dhabi, UAE.}
\thanks{Jingyu Deng is with Multimedia and Visual Computing Lab, NYU Tandon and Abu Dhabi, and Department of Electrical and Computer Engineering, Abu Dhabi, UAE.}
\thanks{* Equal contribution}
\thanks{Corresponding author: Yi Fang(yfang@nyu.edu).}
}



\maketitle

\begin{abstract}
In this paper, we deal with the problem of object detection on remote sensing images. Previous methods have developed numerous deep CNN-based methods for object detection on remote sensing images and the report remarkable achievements in detection performance and efficiency. However, current CNN-based methods mostly require a large number of annotated samples to train deep neural networks and tend to have limited generalization abilities for unseen object categories. In this paper, we introduce a few-shot learning-based method for object detection on remote sensing images where only a few annotated samples are provided for the unseen object categories. More specifically, our model contains three main components: a meta feature extractor that learns to extract feature representations from input images, a reweighting module that learn to adaptively assign different weights for each feature representation from the support images, and a bounding box prediction module that carries out object detection on the reweighted feature maps. We build our few-shot object detection model upon YOLOv3 architecture and develop a multi-scale object detection framework. Experiments on two benchmark datasets demonstrate that with only a few annotated samples our model can still achieve a satisfying detection performance on remote sensing images and the performance of our model is significantly better than the well-established baseline models.
\end{abstract}


\section{Introduction}

\begin{figure*}[ht]
\centering
\includegraphics[width=5.5in]{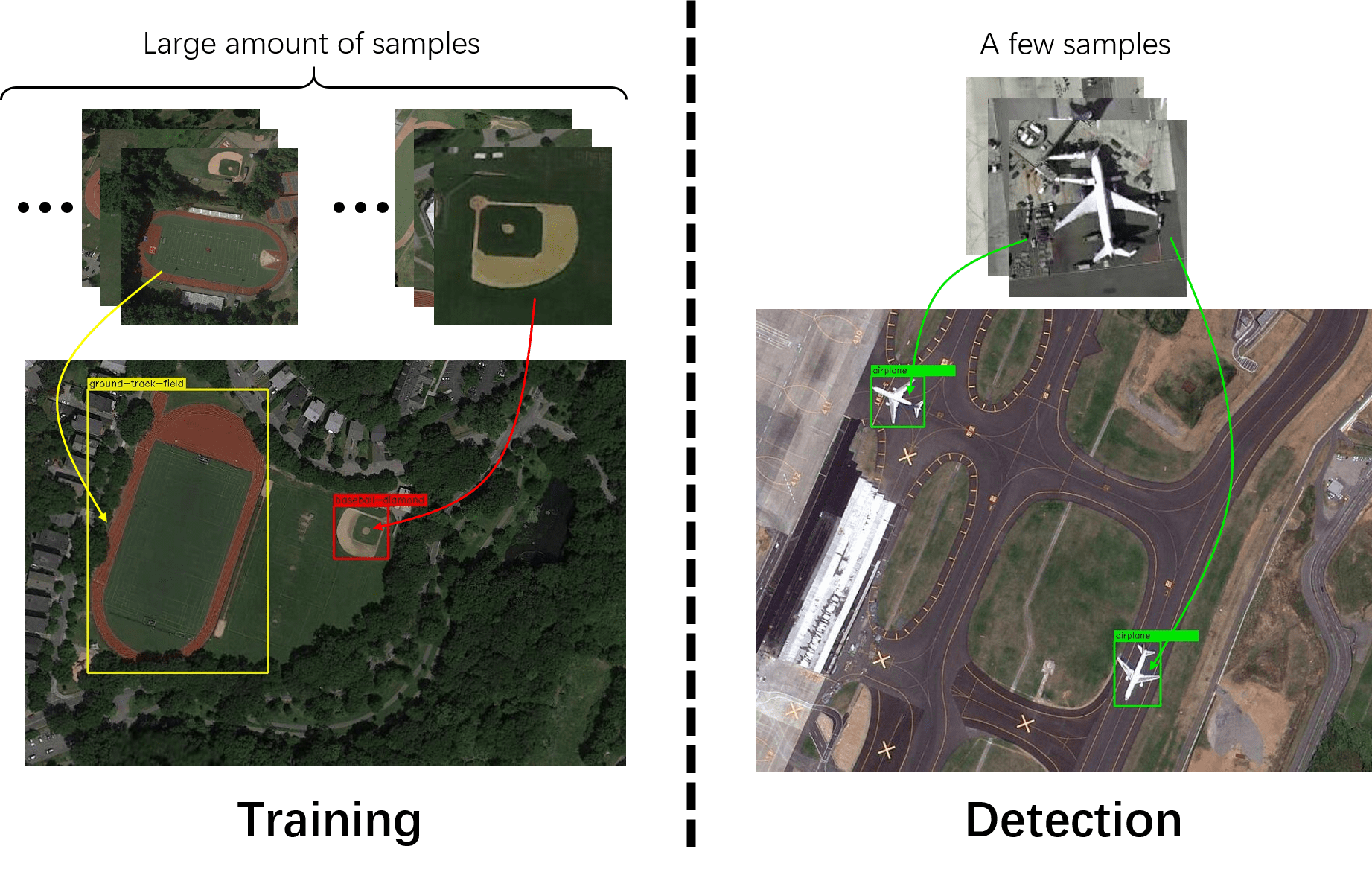}
\caption{Illustration of few-shot detection on remote sensing images. Our model is trained with large number of annotated samples from the base classes and performs detection on unseen classes with only a few annotated samples.}
\label{fig:few_shot}
\end{figure*}

Object detection has been a long-standing problem in both remote sensing and computer vision fields. It is generally defined as identifying the location of target objects in the input image as well as recognize the object categories. Automatic object detection has been widely used in many real-world applications, such as hazard detection, environmental monitoring, change detection, urban planning, etc \cite{cheng2016survey,li2020object}.

In the past decades, object detection has been extensively studied and a large number of methods have been developed for the detection of both artificial objects (e.g. vehicles, buildings, roads, bridges, etc) and natural objects (e.g. lakes, coasts, forests, etc) in remote sensing images. Existing object detection methods in RSIs can be roughly divided into four categories: 1) template matching-based methods, 2) knowledge-based methods, 3) object-based image analysis (OBIA)-based methods and 4) machine learning-based methods \cite{cheng2016survey}. Among them, the machine learning-based methods have powerful abilities for robust feature extraction and object classification and are extensively studied by many recent approaches to achieve significant progress for this problem \cite{bai2014vhr,bi2012visual,huang2009road,volpi2013multi}.

In recent years, among all machine learning-based methods for object detection, deep learning methods, especially convolutional neural networks (CNNs), have drawn huge research attention. Thanks to the powerful feature extraction abilities of CNN models, a huge amount of CNN-based methods have been developed for object detection in both optical images and remote sensing images. Notable methods include Faster R-CNN \cite{ren2015faster}, YOLO\cite{redmon2016you}, SSD \cite{liu2016ssd}. In the remote sensing field, recent works mostly build their methods upon the prevalent architectures in the computer vision field.

Despite the breakthrough achieved by deep learning-based methods for object detection, these methods suffer from a common issue: a large-scale, diverse dataset is required to train a deep neural network model. Any adjustment on the candidate identifiable classes will be expensive for existing methods because collecting a new RSI dataset with a large number of manual annotations is costly and these methods need a lot of time to re-train their parameters on the newly collected dataset. On the other hand, training a model with only a few samples from the new classes tend to suffer from the overfitting problem and the generalization abilities will be greatly reduced. Therefore, a special mechanism of learning robust features from a few samples of the new classes is desired for object detection in RSIs.

In the past few years, few-shot learning has been extensively studied in computer vision field for the task of scene classification \cite{vinyals2016matching,snell2017prototypical,gidaris2018dynamic}, image segmentation \cite{zhang2019canet,wang2019panet,hu2019attention} and object detection \cite{dixit2017aga,karlinsky2019repmet,wang2019few,kang2019few}. Few-shot learning aims at learning to learn transferable knowledge that can be well generalized to new classes and therefore performs image recognition (e.g., classification, segmentation) on new classes with only a few annotated examples. Existing few-shot object detection methods are designed for common objects (e.g. bicycles, cars, chairs, etc) in optical images. These objects are generally of consistent size. While in remote sensing images, objects can have very different sizes and the spatial resolution of RSIs can be quite different, which makes the problem even more challenging when only a few annotated samples are provided. 

In this paper, we introduce a few-shot learning-based method for object detection on remote sensing images. Under the few-shot scenario, our model aims to learn a detection model from the dataset of base classes that can conduct accurate object detection for unseen (novel) classes with only a few annotated samples. Fig. \ref{fig:few_shot} illustrates the basic idea of few-shot object detection on remote sensing images. We build our method upon a recently published paper \cite{kang2019few} which is designed for common object detection in optical images. To address the scale variations that inherent present in remote sensing images, we extend \cite{kang2019few} to a multi-scale feature extraction and object detection framework. Concretely, a meta feature extractor is designed to learn to extract feature representations from input images; a feature reweighting module is designed to learn to adaptively assign different weights for each feature representation from the support images. A bounding-box prediction module carries out object detection on the reweighted feature maps. 
Our few-shot detection method includes two stages: The training stage and the detection stage. In the training stage, our model is trained on a large amount of data from the base classes set and learns to learn meta-knowledge for object detection. In the detection stage, a few samples from the novel classes (no overlapping with the base classes) are used to finetune the model to make it adapted to the novel classes while still maintaining the meta-knowledge during the training stage.

The main contributions of this paper are summarized as follows:
\begin{itemize}
\item In this paper, we introduce the first few-shot learning-based method for object detection on remote sensing images. Our method is trained with large-scale data from some base classes and can learn meta-knowledge from base classes and generalize well to novel classes with only a few samples. 
\item Our method contains three main components: a meta-feature extraction network, a feature reweighting module, and a bounding box prediction module. All three modules are designed in multi-scale architecture to enable multi-scale object detection.
\item Experiments on two public benchmark datasets demonstrate the effectiveness of the proposed method for few-shot object detection on remote sensing images.
\end{itemize}

\section{Related Work}
\subsection{Object detection in Computer Vision}
Object detection is a hot topic in the computer vision field with extensive studies, especially since the boom of deep learning methods. R-CNN \cite{girshick2014rich} is one of the earliest and successful methods that adopt CNN for object detection. In R-CNN, the authors replace the traditional hand-crafted feature engineering process with CNN-based feature learning and demonstrate a significant performance boost. Following R-CNN, Fast R-CNN \cite{girshick2015fast} performs feature extraction on the original input images and map all region proposals onto the extracted feature map. A region of interest (RoI) pooling layer is proposed to transform feature representations of each ROI into a fixed-length vector. Besides, to facilitate neural network design, the SVM classifier is replaced with Softmax classifier and the bounding box regression process is included in the model instead of doing it afterward. Fast R-CNN improves detection efficiency by a large margin. Another important variant comes from Faster R-CNN \cite{ren2015faster}. To further overcome the computation burden from the region proposal generation process, Faster R-CNN introduces a region proposal network (RPN) to generate region proposals from the CNN network and enables weight sharing between the RPN network and detection network. The following works, such as  \cite{lin2017feature, he2017mask}, mostly base their method on Faster R-CNN architecture. For example, Mask R-CNN \cite{he2017mask} adopts Feature Pyramid Network (FPN) \cite{lin2017feature} as the backbone network to produce multi-scale feature maps and adds a mask prediction branch to detect precise boundaries of each instance. 

The aforementioned approaches generally divide the detection process into two stages: region proposal generation, object detection from the region proposals. These methods are therefore often called two-stage object detectors. Another family of methods remove the region proposal generation process and directly conduct object detection on the input images. These methods are therefore often called one-stage object detectors. One of the most successfully one-stage object detectors is YOLO \cite{redmon2016you}. In the YOLO model, the input image is divided into grid cells and each cell is responsible for detecting a fixed number of objects. A deep CNN architecture is designed to learn high-level feature representation for each cell, followed by a successive of fully connected layers to predict the object categories and locations. YOLO is a lot faster than two-stage object detectors but with inferior detection performance. Following variants, YOLOv2 \cite{redmon2017yolo9000} and YOLOv3 \cite{redmon2018yolov3} improve the performance by using more powerful backbone network and conduct object detection on multiple scales. More specifically, the YOLOv3 model adopts FPN \cite{lin2017feature} as the backbone network thus enables more powerfully feature extraction and detection at different scales. Following works mostly improve the performance by using deconvolutional layers \cite{fu2017dssd}, multi-scale detection pipeline \cite{liu2016ssd}, or focal loss \cite{lin2017focal}.

\subsection{Object detection in RSIs} 
Existing methods for object detection on remote sensing images fall into four categories: template matching-based methods, knowledge-based methods, object-based image analysis (OBIA)-based methods, and machine learning-based methods \cite{cheng2016survey}. The template matching-based methods use the stored templates, which are generated through hand-crafting or training, to find the best matches at each possible location in the source image. Typical template matching-based methods include rigid template matching \cite{chaudhuri2012semi,mckeown1988cooperative,zhou2006road} and deformable template matching \cite{fischler1973representation}. Knowledge-based methods treat the object detection problem as a hypothesis testing process by using pre-established knowledge and rules. Two kinds of well-known knowledge are geometric knowledge \cite{huertas1988detecting,mcglone1994projective,trinder1998automatic,weidner1995towards} and context knowledge \cite{akccay2010building,huertas1988detecting,irvin1989methods}. OBIA-based methods start with segmenting images into homogeneous regions that represent a relatively homogeneous group of pixels and then perform region classification using region-level features from hand-crafted feature engineering. The last family of methods, machine learning-based object detectors contains two fundamental processes: hand-crafted feature extraction and classification using machine learning-based algorithms. Machine learning-based methods have shown more powerful generalization abilities compared to the other three families of methods \cite{cheng2016survey}.

Among all machine learning-based methods, deep learning-based methods have drawn huge research attention and are widely used in recent RSI object detection works. Unlike traditional machine learning-based methods that use hand-crafted features, deep learning-based methods use deep neural networks to automatically learn robust features from input images. In this direction, early efforts adopt R‐CNN architecture to detect geospatial objects on remote sensing images \cite{cheng2016learning,deng2017toward,tang2017vehicle,yang2017m,li2017rotation,zhong2018multi,guo2018geospatial,yang2018aircraft}. For example, \cite{cheng2016learning} introduces a new rotation-invariant layer to the R-CNN architecture to enhance the performance for detection objects with different orientations. Following the great success of Faster R-CNN, numerous works have tried to extend the Faster R-CNN framework to remote sensing community \cite{zou2016ship,lin2017fully,liu2018arbitrary,tang2017arbitrary}. For example, \cite{li2017rotation} develops a rotation-insensitive RPN by using multi-angle anchors instead of horizontal anchors used in conventional RPN network. The proposed method can effectively detect geospatial objects of arbitrary orientations.

Following the great success of one-stage based methods for object detection on natural images, researches also developed various regression-based methods for object detection on remote sensing images \cite{tang2017arbitrary,liu2017learning,tang2017vehicle,hu2019sample}. For example, \cite{tang2017arbitrary} extends SSD model to conduct real-time vehicle detection on remote sensing images. \cite{liu2017learning} replaces the horizontal anchors with oriented anchors in SSD \cite{liu2016ssd} framework, thus enables the model to detect objects with orientation angles. Following methods further enhance the performance of geospatial object detection on remote sensing images by using hard example mining \cite{tang2017vehicle}, multi‐feature fusion \cite{zhong2017robust},  transfer learning \cite{han2017efficient}, non‐maximum suppression \cite{xu2017deformable}, etc.

\subsection{Few-shot detection} 
Few-shot learning, as one of the supervised meta-learning methods, aims at learning to learn transferable knowledge that can be generalized to new classes and therefore performs image recognition (e.g., classification, detection, segmentation) on new classes when only a few annotated samples are given. In recent years, few-shot detection is receiving growing attention recently in the computer vision field. \cite{chen2018lstd} proposes to fine-tune a pre-trained model, such as Faster R-CNN \cite{ren2015faster} and SSD \cite{liu2016ssd}, on few given examples to transfer it into a few-shot detector. In \cite{dong2018few}, the authors enrich the training examples with additional unannotated data in a semi-supervised setting and obtain performance comparable to weakly supervised methods with a large amount of training data.  \cite{karlinsky2019repmet} introduce a Distance Metric Learning sub-net to replace the classification head of standard detection architecture \cite{lin2017feature}, and achieves satisfying detection performance with a few training samples. \cite{kang2019few} introduces a reweighting module to produce a group of reweighting vectors from a few supporting samples, one for each class, to reweight the meta feature extracted from the DarkNet-19 network. With the reweighted meta-features, a bounding box prediction module is adopted to produce the detection results. However, the DarkNet-19 only produces a single meta feature map for each input image, leading to poor performance when detecting objects with large size variations. In contrast to \cite{kang2019few} that only conducts object detection on a single scale feature map, our proposed method extracts hierarchy feature maps with different scales from an FPN-like structure and improves the performance by performing multi-scale object detection under the few-shot scenario.

\section{Method}

\begin{figure*}[ht]
\centering
\includegraphics[scale=0.34]{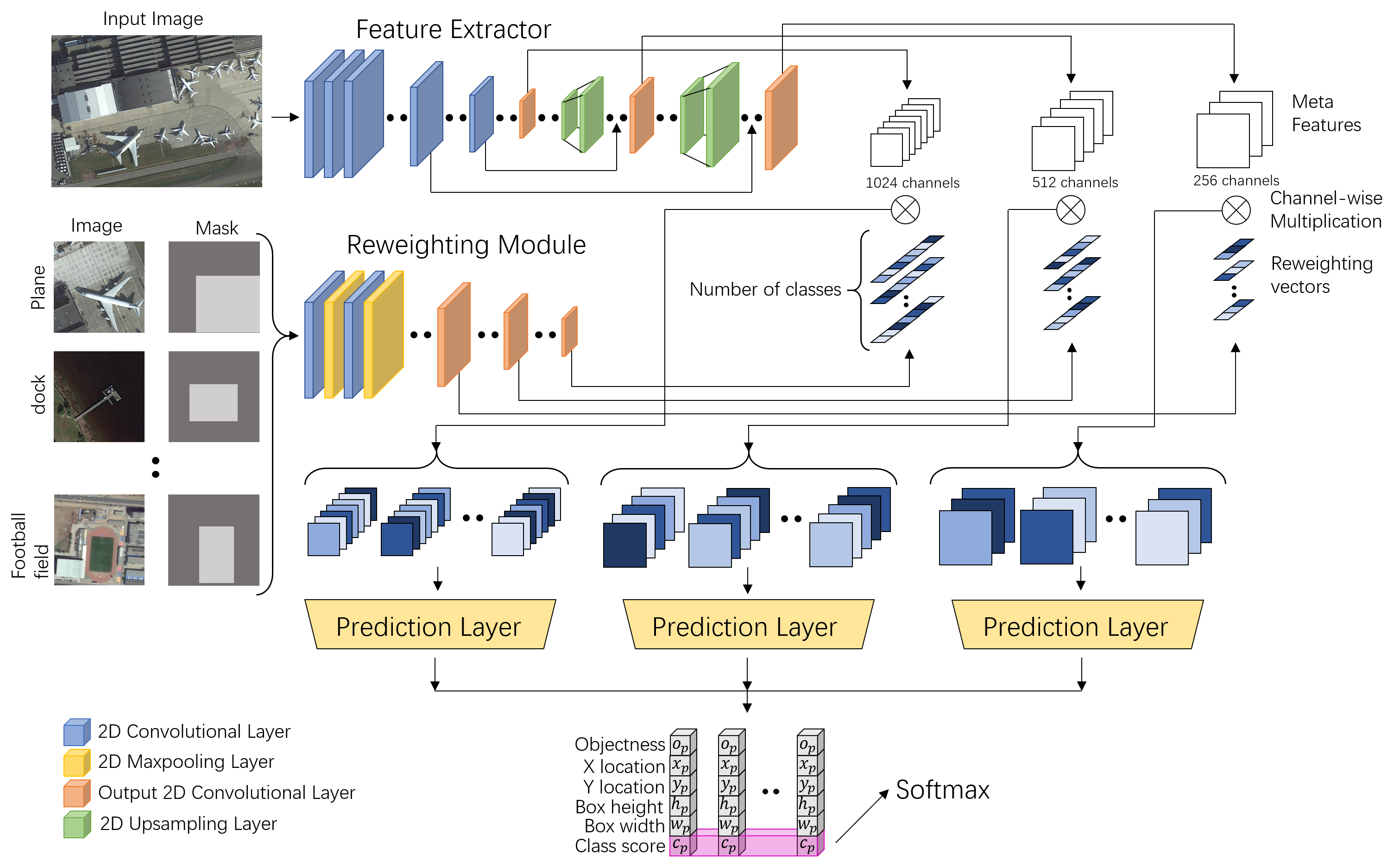}
\caption{The pipeline of the proposed method for few-shot object detection on remote sensing images. Our method consists of three main components: a Meta Feature Extractor, a Reweighting Module, and a Bounding Box Prediction Module. The Feature Extractor network takes a query image as input and produces meta feature maps at three different scales. The Reweighting Module takes as input $N$ support images with labels, one for each class, and outputs three groups of $N$ reweighting vectors. These reweighting vectors are used to recalibrate the meta-feature maps of the same scale through a channel-wise multiplication. The reweighted feature maps are then fed into three independent bounding box detection modules to predict the objectness scores ($o$), the bounding box locations and sizes ($x$, $y$, $w$, $h$) and class scores ($c$) at three different scales.}
\label{fig:pipeline}
\end{figure*}

\subsection{Method Overview}
We first clarify the settings for the few-shot object detection problem. The problem of few-shot object detection aims at learning a detection model from the dataset of available classes (base classes) that can conduct object detection on images from unseen classes (novel classes) with only a few annotated samples from the same unseen classes. For each base class, there are adequate samples for model training. While the novel classes have only a few annotated samples. A few-shot object detection model should be able to learn meta-knowledge from the dataset of base classes and well transfer it to the novel classes.

This few-shot object detection setting is very common in real-world scenarios--one may need to develop a new object detection model while collecting a large scale dataset for the target classes is time-consuming. A good start point would be deploying a detection model pre-trained on some existing large-scale object detection datasets (e.g., DIOR \cite{li2020object}). However, these datasets only cover a limited number of object categories, while one may only focus on several specific object categories that may not happen to be included in these datasets.

To facilitate model training and evaluation, we construct several episodes from the training and testing set. Each episode $E_i$ is constructed from a set of support images $S_i$ (with annotations) and a set of query images $Q_i$. Given a K-shot segmentation task, each support set $S_i$ consists of $K$ annotated images per object category. We denote the support set as, $S_i = \{(I_{k}, M_{k})\}$ where $I_{k}$ denotes the input image and $I_{k} \in \mathbb{R}^{h \times w \times 3}$, and $M_{k}$ denotes the bounding box annotation, $k=1,2,...K$. The query set $Q_i$ contains $N_q$ images from the same set of class $\mathcal{C}_i$ as the support set. The support images are used for meta knowledge learning and our model performs object detection for the query images by applying the learned meta-knowledge on support images. 

Figure \ref{fig:pipeline} illustrates the pipeline of the proposed method. Our few-shot object detection model (FSODM) is designed to leverage the meta-knowledge from the dataset of base classes. To achieve this goal, a Meta Feature Extractor module is first developed to learn meta-features at three different scales from input query images. Then a Reweighting Module takes as input $N$ support images with labels, one for each class, and outputs three groups of $N$ reweighting vectors, one for each scale. These reweighting vectors are used to recalibrate the meta-features of the same scale through a channel-wise multiplication. With the reweighting module, the meta-information from support samples is extracted and used to amplify those meta-features that are informative for detecting novel objects in the query images. The reweighted meta-features are then fed into three independent bounding box detection modules to predict the objectness scores ($o$), the bounding box locations, and sizes ($x$, $y$, $w$, $h$) and class scores ($c$) at three different scales.

\subsection{Meta Feature Extractor}

Our meta feature extractor network is designed to extract robust feature representations from input query images. Unlike \cite{kang2019few} that only extract single scale meta features, objects in remote sensing images can have quite different sizes. Therefore, a multi-scale feature extraction network is desired. In this paper, our feature extractor network is designed based on DarkNet-53 \cite{redmon2018yolov3} and FPN. The detailed network architecture can be found in \cite{redmon2018yolov3}. For each input query image, our meta feature extractor network produces meta features at three different scales. Let $I$ be the input query image, $I \in \mathbb{R}^{h \times w \times c}$, the generated meta features after the feature extractor network can be formulated as:
\begin{equation}
F_{i}=\mathcal{E}(I) \in \mathbb{R}^{h_i \times w_i\times m_i}
\end{equation}
where $i$ denotes the scale level, $i \in {1,2,3}$, $h_i, w_i$ and $m_i$ denote the size of feature map at scale $i$.

In this paper, we choose feature maps at the scales of 1/32x, 1/16x and 1/8x, i.e., the output feature maps will have sizes of ($h/32 \times w/32 \times 1024$), ($h/16 \times w/16 \times 512$) and ($h/8 \times w/8 \times 256$). 

\subsection{Feature Reweighting Module}
\label{Feature Reweighting}
Our feature reweighting module is designed to extract meta-knowledge from the support images. To achieve this goal, a light-weight CNN is formulated to map each support image to a set of reweighting vectors, one for each scale. These reweighting vectors will be used to adjust the contribution of meta-features and highlight meta-features significant for novel objects detecting.

Assuming the support samples are from $N$ object categories, our feature reweighting module receives inputs of $N$ support images and their masks. For each one of the $N$ classes, one support image $\widehat{I}_{j}$ along with its corresponding bounding box annotations $M_{j}$ will be randomly chosen from the support set. Then our feature reweight module maps it into a class-specific representation $V_{ij} \in\mathbb{R}^{m_i}$ with $V_{ij} = \mathcal{M}(\widehat{I}_{j}, M_j)$. The reweighting vector $V_{ij}$ will be used to reweight the meta-features and highlight the informative one at scale $i$ and class $j$. 

Table \ref{tab:reweight} shows the network architecture of our feature reweighting module $\mathcal{M}$. The output reweighting vectors are taken from every Global Maxpooling layers (marked with an underline in Table \ref{tab:reweight}) and each reweighting vector has the same dimension as the corresponding meta-feature. 
After obtaining the meta features $F_{i}$ and the reweighting vectors $V_{ij}$, we compute the class-specific reweighted feature maps $\widehat{F}_{ij}$ by:
\begin{equation}
\widehat{F}_{ij}=F_{i} \otimes V_{ij},\ \ i=1,2,3\ and\ j=1,2,...,N
\end{equation}
where $\otimes$ is channel-wise multiplication which is realized through $1 \times 1$ convolution with the reweighting vectors $V_{ij}$ as the convolution kernels.

As we can see, after channel-wise multiplication, there will be three groups of reweighted feature maps, one for each scale. In each group, our feature reweighting module produces $N$ reweighted feature maps. Each reweighted feature map is responsible for detecting objecting objects at one of the $N$ class.

\begin{table}[h]
    \centering
    \begin{tabular}{cccccc}
    \toprule
    Index&Type&Filters&Size&Output\\
    \midrule
    1&Convolutional&32&3$\times$3/1&512$\times$512\\
    2&Maxpooling&&2$\times$2/2&256$\times$256\\
    3&Convolutional&64&3$\times$3/1&256$\times$256\\
    4&Maxpooling&&2$\times$2/2&128$\times$128\\
    5&Convolutional&128&3$\times$3/1&128$\times$128\\
    6&Maxpooling&&2$\times$2/2&64$\times$64\\
    7&Convolutional&256&3$\times$3/1&64$\times$64\\
    8&Maxpooling&&2$\times$2/2&32$\times$32\\
    9&Convolutional&256&3$\times$3/1&32$\times$32\\
    10&\underline{GlobalMax}&&32$\times$32/1&1$\times$1\\
    11&Route&8\\
    12&Convolutional&512&3$\times$3/1&32$\times$32\\
    13&Maxpooling&&2$\times$2/2&16$\times$16\\
    14&Convolutional&512&3$\times$3/1&16$\times$16\\
    15&\underline{GlobalMax}&&16$\times$16/1&1$\times$1\\
    16&Route&13\\
    17&Convolutional&1024&3$\times$3/1&16$\times$16\\
    18&Maxpooling&&2$\times$2/2&8$\times$8\\
    19&Convolutional&1024&3$\times$3/1&8$\times$8\\
    20&\underline{GlobalMax}&&8$\times$8/1&1$\times$1\\
    \bottomrule
    \end{tabular}
    \vspace*{0.1in}
    \caption{Network architecture of the reweighting module . 'Convolutional' denotes 2D Convolutional Layer. 'Filters' is the number of convolutional filters. 'Size' indicates the convolutional kernel size and stride of the layer in the form of 'kernel height $\times$ kernel width / stride'; 'Maxpooling' denotes 2D Maxpooling Layer; 'GlobalMax' denotes 2D Global Maxpooling Layer where with kernel size equals to input size; 'Route' is a layer that used to control data route. The 'Route' layer will take the output of layer in 'Filter' column as the input of the next layer. For example, the 'Route 8' in Layer 11 means taking the output of Layer 8 as the input of next layer, i.e., Layer 12.}
    \label{tab:reweight}
\end{table}

\subsection{Bounding Box Prediction}

\begin{figure*}[t]
    \centering
    
    \subfigure[]{
        \includegraphics[width=2.2in]{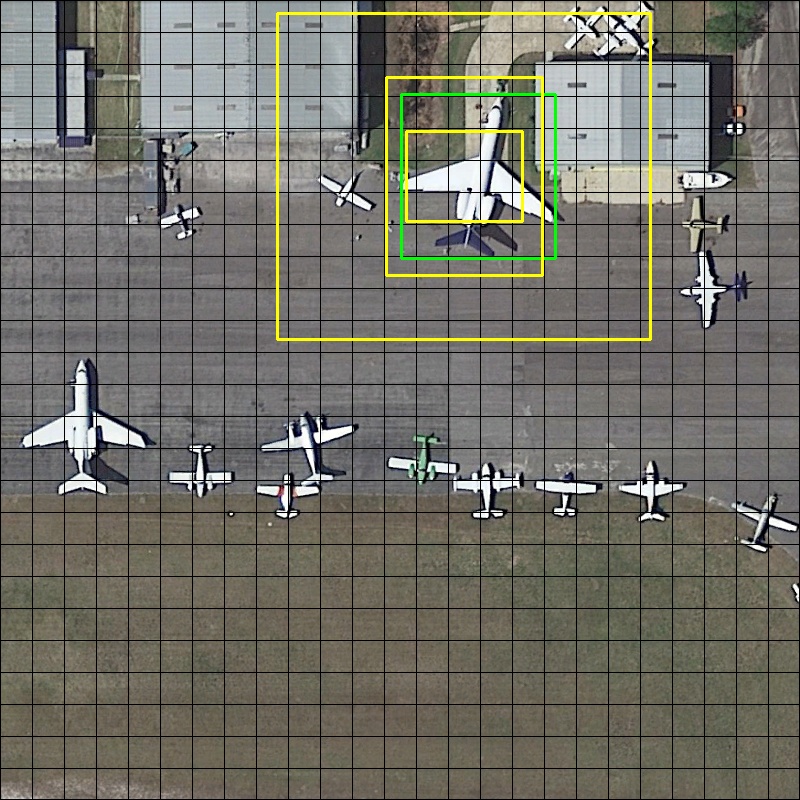}
        \label{fig:anchor_1}
    }
    \hspace{0in}
    \subfigure[]{
        \includegraphics[width=2.2in]{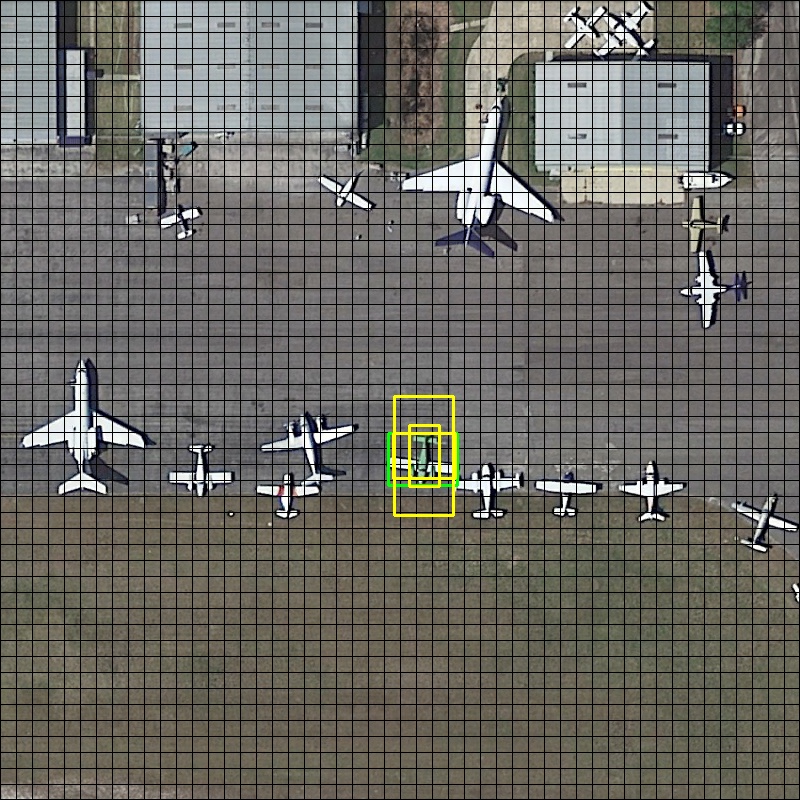}
        \label{fig:anchor_2}
    }
    \hspace{0in} 
    \subfigure[]{
        \includegraphics[width=2.2in]{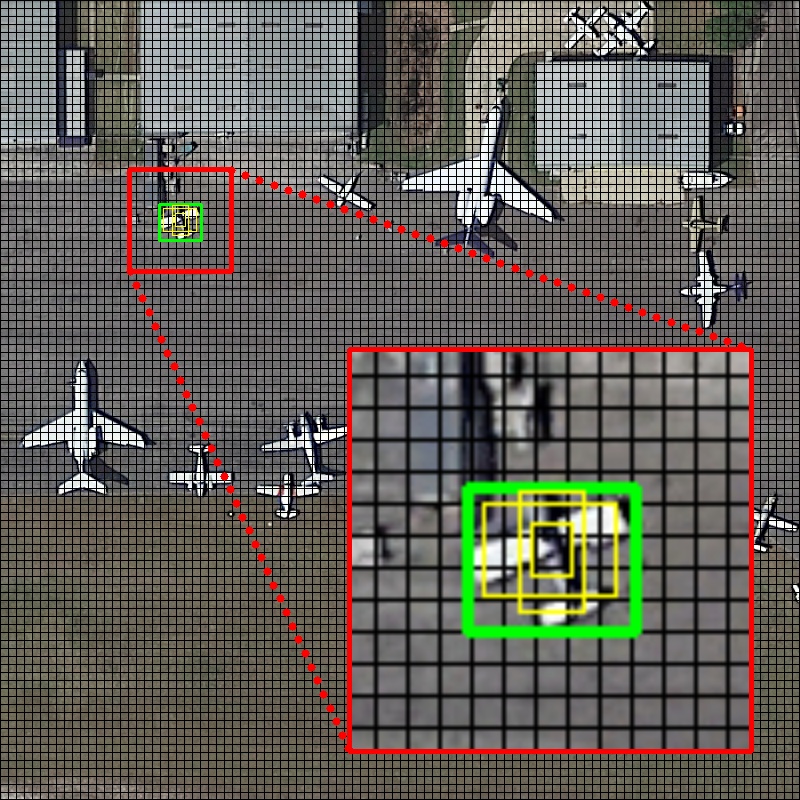}
        \label{fig:anchor_3}
    }
    \caption{Anchor boxes on three feature maps of different scales. Green box is a ground truth bounding box of the object, and yellow boxes are anchor boxes of the prediction cell. The size of the input image is 800$\times$800, and (a), (b) and (c) are anchor settings of its small (25$\times$25), middle (50$\times$50) and large (100$\times$100) feature maps.}
    \label{fig:anchor}
\end{figure*}

\begin{figure}[h]
\centering
\includegraphics[scale=0.5]{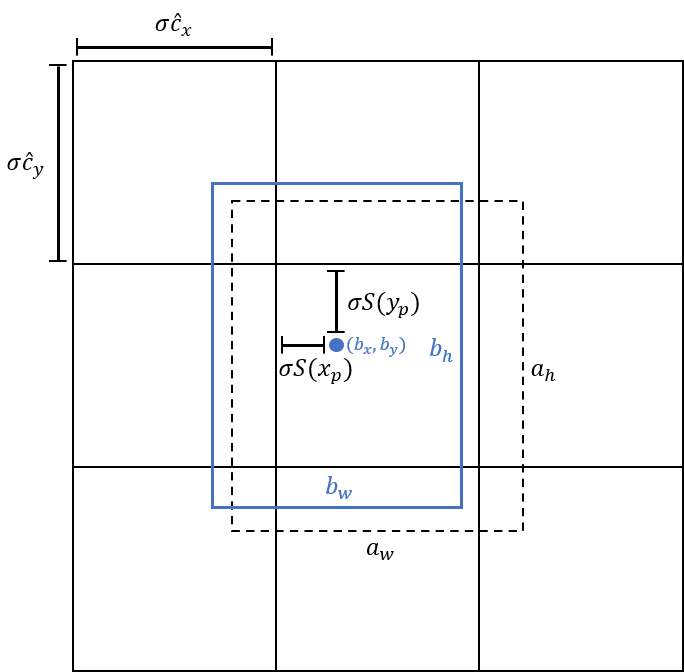}
\caption{Illustration of anchor boxes and predicted bounding box representations. The solid line grid is cells of feature map, the dash line rectangle is a anchor box and the blue line rectangle is a predicted box.}
\label{fig:bbox}
\end{figure}

Our bounding prediction module ($\mathcal{P}$) takes as input the reweighted feature maps and produces the object categories and bounding locations. Following the setting of YOLOv3 \cite{redmon2018yolov3}, at each scale, we predict three bounding boxes for each of the class-specific features maps. To achieve this goal, we generate a set of anchor boxes at each pixel location on the input feature maps, Fig. \ref{fig:anchor} illustrate the anchor box settings at three different scales. For the first feature map with scale level $i$ equals 1, the size of anchor boxes are set to ($116 \times 90$), ($156 \times 198$), ($373 \times 326$). For the second feature map with scale level $i$ equals 2, the size of anchor boxes are set to ($30 \times 61$), ($62 \times 45$), ($59 \times 119$) for the middle feature map. For the third feature map with scale level $i$ equals 3, the size of anchor boxes are set to ($10 \times 13$), ($16 \times 30$), ($33 \times 23$).

For each anchor box in the feature map, our bounding box prediction module produces a 6-dimensional output as displayed in Fig. \ref{fig:pipeline}. Among the output, the first 4 elements are used for object location prediction, and the left 2 elements are the objectness score ($o_{p}$) and classification score ($c_{p}$).
Fig. \ref{fig:bbox} shows the output representation of each bounding box. Assuming the coordinates of a predicted bounding box are $b_{x}$, $b_{y}$, $b_{w}$, $b_{h}$, where $b_x$ and $b_y$ are the coordinates of its center, $b_w$ and $b_h$ are the width and height of the bounding box. Instead of directly regress the bounding box locations, our bounding box prediction module predicts 4 offset values $x_{p}$, $y_{p}$, $w_{p}$, $h_{p}$ and coordinates of the predicted box can be computed through:
\begin{equation}
\begin{aligned}
b_{x}&=\sigma(S(x_{p})+\widehat{c}_{x})\\
b_{y}&=\sigma(S(y_{p})+\widehat{c}_{y})\\
b_{w}&=a_{w}e^{w_{p}}\\
b_{h}&=a_{h}e^{h_{p}}
\end{aligned}
\label{equ:bbox}
\end{equation}
where $S(x)$ is the Sigmoid function and $\sigma$ is a scale transformation coefficient equivalents to the ratio between the input image side length and the feature map side length; $\widehat{c}_{x}$ and $\widehat{c}_{y}$ are cell offsets from the top left corner to the cell that makes prediction; $a_{w}$ and $a_{h}$ are the width and height of the corresponding anchor box. 


The objectness score ($o_{p}$) implies the possibility of the existences of an object which can be computed as $P_{o}=S(o_{p})$, where $P_{o}$ is objectness possibility and $S(\cdot)$ is the sigmoid function. Cause we have one set of reweighted feature maps for each class, each predicted bounding box only needs one score for class prediction instead of the total number of categories ($N$). The classification score ($c_{p}$) indicates the possibility that the detected object belongs to each one of the classes. Taking the classification scores generated from the same anchor boxes locations with the same anchor sizes as a group, there will be $N$ classification scores belong to the same anchor boxes of the input image. Naming these $N$ predicted boxes as $c_{pi}$ ($i=1,2,...,N$), a softmax function is applied on the probability vector to normalize these probability values. The final classification score for each class $i$ can be formulated as:
\begin{equation}
P_{ci}=\frac{e^{c_{pi}}}{\sum_{j=1}^{N}e^{c_{pj}}}
\end{equation}
$P_{ci}$ is the final classification possibility of class $i$ and $\sum_{i=1}^{N}P_{ci}=1$. Objectness possibility and classification possibility together can help to judge whether an object is detected and which class the object belongs to.

\subsection{Loss function}
The loss function of our object detection model contains two parts, object localization loss and object classification loss. For object localization, we use the mean-square-error loss to penalize the misalignment between the predicted bounding boxes and the ground truth ones. Given the predicted bounding boxes coordinates $coord_{p}$ and ground truth bounding boxes coordinates $coord_{t}$, the object localization loss is calculated as:
\begin{equation}
\begin{aligned}
\mathcal{L}_{loc}&=\frac{1}{N_{pos}}\sum_{pos} \sum_{l}(coord_{t}^{l}-coord_{p}^{l})^2
\end{aligned}
\end{equation}
where $l$ denotes a coordinate enumerator, it can be chosen from $\{w,y,w,h\}$, i.e., the four coordinate representation of a specific bounding box. $pos$ indicates all positive anchors which are expected to predict a ground truth bounding box. Only losses of positive anchors are used in coordinate loss computing and localization losses of those negative anchor boxes are ignored. We identify an anchor box as positive if the IoU between this anchor box with a certain ground truth bounding box is larger than a given threshold (e.g., 0.7). Also, we identify an anchor box as negative if the IoU between this anchor box with all ground truth bounding box is less than a given threshold (e.g., 0.3). We also identify an anchor box as positive if it has the largest IoU with a certain ground truth bounding box among all anchor boxes.

The loss function for objectness score $\mathcal{L}_o$ is binary cross-entropy loss, calculated as:
\begin{equation}
\begin{aligned}
\mathcal{L}_{obj}&=\frac{1}{N_{pos}}\sum_{pos} -[P_{t}\cdot\log{P_{o}}+(1-P_{t})\cdot\log{(1-P_{o})}]\\
&=\frac{1}{N_{pos}}\sum_{pos} -\log{P_{o}}\\
\mathcal{L}_{noobj}&=\frac{1}{N_{neg}}\sum_{neg} -[P_{t}\cdot\log{P_{o}}+(1-P_{t})\cdot\log{(1-P_{o})}]\\
&=\frac{1}{N_{neg}}\sum_{neg} -\log{(1-P_{o})}\\
\mathcal{L}_{o}&=w_{obj}\cdot\mathcal{L}_{obj}+w_{noobj}\cdot\mathcal{L}_{noobj}\\
\end{aligned}
\end{equation}
, where $P_{o}$ denotes the predicted objectness possibility mentioned above; $P_{t}$ denotes the true possibility which is one when it is a positive box and is zero when negative; $w_{obj}$ and $w_{noobj}$ are weights of objectness loss and none-objectness loss. Considering there are usually a lot more negative boxes than positive boxes, $w_{obj}$ and $w_{noobj}$ are used to balance these two loss terms. 

For object classification, we use cross-entropy loss to enforce the predicted classes to be align with the ground truth ones, calculated as:
\begin{equation}
\begin{aligned}
\mathcal{L}_{c}=\frac{1}{N_{pos}}\sum_{pos} -\log{(\frac{e^{c_{pt}}}{\sum_{j=1}^{N}e^{c_{pj}}})}
\end{aligned}
\end{equation}
, where $c_{pt}$ is the classification score of the true class. Because we already use objectness score to decide whether the predicted box contains an object or not, so the background class is ignored during classification loss calculation. The overall objective loss function is formulated as: 
\begin{equation}
\mathcal{L}=\mathcal{L}_{loc} + \mathcal{L}_o+\mathcal{L}_c
\label{eq_loss_all}
\end{equation}

\subsection{Training and Inference}
In our few-shot detection model, training process is conducted on episodes. To facilitate model train in few-shot detection scenerio, during training, we reorganized the training dataset into two sets: query set ($\mathcal{Q}$) and support set ($\mathcal{S}$). Query set contains a set of query images and their annotations ($A$):
\begin{equation}
\begin{aligned}
\mathcal{Q}=\{(I,A)\}
\end{aligned}
\end{equation}
Support set is training dataset regrouped by object classes. As explained in \ref{Feature Reweighting}, each query image is associated with a group of support images from all classes. Therefore, we separate training images into $N$ groups $\widehat{I}_i, i=1,2,...,N$, according to object categories in those images. After regrouping, a bounding box mask is generated for each support image. The mask is generated by setting the pixel value to 1 when the pixel located within the ground truth bounding box and 0 otherwise. Assuming $M_{\widehat{I}}$ is the bounding box mask of $\widehat{I}$, Support set can be formulated as:
\begin{equation}
\begin{aligned}
\mathcal{S}=\{(\widehat{I}_1, M_{\widehat{I}_1}),(\widehat{I}_2, M_{\widehat{I}_2}),...,(\widehat{I}_N, M_{\widehat{I}_N})\}
\end{aligned}
\end{equation}
Each episode $\mathcal{T}_j$ consists of one query image $I_j$, the bounding box annotation $A_j$ of query image, and the pair of one support image and its bounding box mask $(\widehat{I}^j_i,M^j_i)$ from each class-specific group:
\begin{equation}
\begin{aligned}
\mathcal{T}_j&=\mathcal{Q}_j\cup\mathcal{S}_j\\
&=\{(I_j,A_j)\}\cup\{(\widehat{I}_{1j},M_{\widehat{I}_{1j}}),(\widehat{I}_{2j},M_{\widehat{I}_{2j}}),...\}
\end{aligned}
\end{equation}
$I_j$ and $\{(\widehat{I}_{ij},M_{\widehat{I}_{ij}})\}$ are inputted into Feature Extractor and \\
Reweighting Module respectively while $A_j$ is used as the ground truth.

Under the few-shot detection scenario, we need to leave some object classes in the dataset for few-shot tuning. To achieve this, all classes in the dataset are divided into base classes and novel classes. Base classes require as many samples as possible to train a precise basic model while novel classes are viewed as a new detection task with only a few annotated samples.

The training process is divided into two steps. The first step is training on base classes to learn the network parameters. This step generally requires a large amount of training data, spends a relatively long time, and usually is not necessary to do it again in following utilization; The second step is turning on novel classes with a few samples, which is fast and is going to be conducted whenever adding new classes.

The overall training and testing process is illustrated in Algorithm \ref{alg1}

\begin{algorithm}[h]
\caption{Training and testing process.}
\begin{algorithmic}[1]
\label{alg1}
\STATE Construct training set $D_{train}$ from base classes and testing set $D_{test}$ from novel classes.
\STATE Initialize the network parameters in the feature extractor network, feature reweighting module, and bounding box prediction module.
\FOR{each training episode $(S_j, Q_j) \in D_{train}$}
\STATE Model training.
\ENDFOR

\FOR{each training episode $(S_j, Q_j) \in D_{test}$}
\STATE Model fine-tuning.
\ENDFOR

\FOR{each testing episode $(S_j, Q_j) \in D_{test}$}
\STATE Extract feature maps for query images using Meta Feature Extractor.
\STATE Generate class-specific reweighting vectors and compute the reweighted feature maps.
\STATE Generate predicted bounding boxes using the bounding box prediction module.
\ENDFOR
\end{algorithmic}
\end{algorithm}

\section{Experiments and Results}\label{Experiments}
In this section, we evaluate the performance of our model for the few-shot object detection on two public benchmark RSI datasets and compare our method with \cite{kang2019few} and \cite{redmon2018yolov3} to show the superiority of our model.

\subsection{Dataset}\label{sc_dataset}
\noindent \textbf{NWPU VHR-10} is a very high resolution (VHR) remote sensing image dataset released by \cite{cheng2014multi}. This dataset contains 800 RSIs collected from Google Earth and ISPRS Vaihingen dataset \cite{niemeyer2014contextual}. 150 "negative samples" without target objects and 650 "positive samples" with at least one object are annotated manually. There are in total 10 object categories in this dataset: airplane, baseball diamond, basketball, bridge, court, ground track field, harbor, ship, storage tank, tennis court, and vehicle.

\noindent \textbf{DIOR} is a large-scale benchmark dataset for object detection on RSIs, released by \cite{li2020object}. Images in the DIOR dataset are collected from GoogleEarth with 23,463 images and 192,472 instances of 20 classes. The object classes include airplane, airport, baseball field, basketball court, bridge, chimney, dam, expressway service area, expressway toll station, harbor, golf course, ground track field, overpass, ship, stadium, storage tank, tennis court, train station, vehicle, and windmill. All images are in the size of 800$\times$800 pixels and the spatial resolutions range from 0.5m to 30m. In the DIOR dataset, the object sizes vary widely. 

\subsection{Experimental Configurations}
To evaluate the detection performance of our FSODM model under the few-shot scenario, we divide each dataset into two parts, one is constructed from the base classes, the other is constructed from the novel classes. For the NWPU VHR-10 dataset, 4 classes (airplane, baseball diamond, and tennis court) are used as novel classes and the others as base classes. For DIOR, 5 classes (airplane, baseball field, tennis court, train station, windmill) are chosen as novel classes and the others as base classes.

Moreover, we apply a multi-scale training technique process to enhance model performance. The scale range of input images varies in (384, 416, 448, 480, 512, 544, 576, 608, 640) and all input images are square. We note that in the DIOR dataset, original images are much larger than the desired input scales. Therefore, those large images are cropped into a series of patches with 1024$\times$1024 pixels and a stride of 512 pixels (For DIOR dataset, all images are with a size of 800$\times$800 pixels, this step is ignored). For the objects get truncated in this process, we ignore these truncated object instances which have an overlapping less than 70\% with the original object instances. 

\subsection{Comparing methods}
We compare our FSODM model with the prevalent object detector YOLOv3 \cite{redmon2018yolov3} model and the current state-of-the-art few-shot detector of \cite{kang2019few}. We do not include other comparing methods because previous works have shown the superiority of YOLOv3 in the standard object detection scenario and \cite{kang2019few} in the few-shot detection scenario. For \cite{kang2019few}, the experimental settings are the same as our method: training on the same set from base classes and tuning on the same set from novel classes. There are some differences when training the YOLOv3 model. The training process of the YOLOv3 model consists of two steps of pre-training and few-shot tuning. In the pre-training stage, we remove all objects belong to novel classes from training data and train the model normally; In the tuning stage, we train the model with a few annotated samples. Note that YOLOv3 model uses complicated data augmentation strategies to enhance its performance, in our experiments, we do not implement these strategies for a fair comparison with our method. 

    
\begin{table*}[h]
    \centering
    \begin{tabular}{p{30mm}<{\centering} | p{10mm}<{\centering} p{10mm}<{\centering} p{10mm}<{\centering} | p{10mm}<{\centering} p{10mm}<{\centering} p{10mm}<{\centering} | p{10mm}<{\centering} p{10mm}<{\centering}}
    \hline
     \multicolumn{1}{c}{}&\multicolumn{3}{c}{FSODM (Ours)}&\multicolumn{3}{c}{\cite{kang2019few}}&\multicolumn{2}{c}{YOLOv3}\\
    \hline
   Class &3-shot&5-shot&10-shot&3-shot&5-shot&10-shot&10-shot&20-shot\\
    \hline
    airplane&0.15&0.58&0.60&0.13&0.24&0.20&0.14&0.30\\
    baseball diamond&0.57&0.84&0.88&0.12&0.39&0.74&0.26&0.50\\
    tennis court&0.25&0.16&0.48&0.11&0.11&0.26&0.01&0.03\\
    \hline
    mean&0.32&0.53&0.65&0.12&0.24&0.40&0.14&0.28\\
    \hline
    \end{tabular}
    \caption{Few-shot detection performance (mAP) on the novel classes of NWPU VHR-10 dataset.}
    \label{tab:few_detect_per_nwpu}
\end{table*}

\begin{table*}[h]
    \centering
    \begin{tabular}{p{30mm}<{\centering} | p{10mm}<{\centering} p{10mm}<{\centering} p{10mm}<{\centering} | p{10mm}<{\centering} p{10mm}<{\centering} p{10mm}<{\centering} | p{10mm}<{\centering} p{10mm}<{\centering} p{10mm}<{\centering}}
    \hline
    \multicolumn{1}{c}{}&\multicolumn{3}{c}{FSODM (Ours)}&\multicolumn{3}{c}{\cite{kang2019few}}&\multicolumn{3}{c}{YOLOv3}\\
    \hline
   Class &5-shot&10-shot&20-shot&5-shot&10-shot&20-shot&10-shot&20-shot&30-shot\\
    \hline
    airplane&0.09&0.16&0.22&0.09&0.15&0.19&0.02&0.07&0.09\\
    baseball field&0.27&0.46&0.50&0.33&0.45&0.52&0.32&0.36&0.45\\
    tennis court&0.57&0.60&0.66&0.47&0.54&0.55&0.29&0.40&0.42\\
    train station&0.11&0.14&0.16&0.09&0.07&0.18&0.01&0.05&0.08\\
    wind mill&0.19&0.24&0.29&0.13&0.18&0.26&0.04&0.12&0.21\\
    \hline
    mean&0.25&0.32&0.36&0.22&0.28&0.34&0.14&0.20&0.25\\
    \hline
    \end{tabular}
    \caption{Few-shot detection performance (mAP) on the novel classes of DIOR dataset.}
    \label{tab:few_detect_per_dior}
\end{table*}

We adopt mean average precision (mAP) to evaluate the object detection performance. We follow the PASCAL VOC2007 benchmark \cite{everingham2007pascal} to calculate mAP which takes the average of 11 precision values when recall increases from 0 to 1 with a step of 0.1.

\subsection{Results on NWPU VHR-10}\label{sc_results_nwpu}
Table \ref{tab:few_detect_per_nwpu} lists the few-shot object detection performance of our FSODM method and the comparing methods on the novel classes of NWPU VHR-10 dataset. As shown in Table \ref{tab:few_detect_per_nwpu}, our proposed FSODM model achieves significantly better performance than \cite{kang2019few} and YOLOv3. More specifically, compared to another few-shot object detector \cite{kang2019few}, our method obtains a mean mAP 166.6\% higher in the 3-shot setting, 120.8\% higher in 5-shot setting, and 62.5\% higher in the 10-shot setting. Conventional none few-shot-based method YOLOv3 obtains a lot worse performance than the two few-shot-based methods. Even under the 20-shot setting, YOLOv3 only gets an mAP of 0.28, which is worse than our FSODM model under the 3-shot setting. Moreover, as shown in Table \ref{tab:few_detect_per_nwpu}, with the increase in the number of annotated samples in novel classes, the detection performance of our FSODM model increases fast. 

From Table \ref{tab:few_detect_per_nwpu} one can also see that both our FSODM model and the comparing methods obtain better performance on the `baseball diamond' category. This is because baseball diamonds have smaller size variations, which makes it to be easily detected by a detection model. 

\subsection{Results on DIOR}\label{sc_results_dior}
Considering the DIOR dataset is a large scale dataset with large variations in object structures and sizes, a larger number of annotated samples are used for the novel classes. Specifically, for the none few-shot-based method, i.e., YOLOv3, we conduct experiments with 20 and 30 annotated samples for the novel classes. Table \ref{tab:few_detect_per_dior} shows the quantitative results of our method and the comparing methods on the novel classes of the DIOR dataset. 

As shown in Table \ref{tab:few_detect_per_dior}, our FSODM model achieves better performance than another few-shot-based method in \cite{kang2019few}. Both these two few-shot-based methods achieve a lot better performance than the none few-shot-based method YOLOv3, even with fewer samples. Moreover, with the increase in the number of annotated samples in novel classes, the detection performance improves consistently for all three methods. Table \ref{tab:few_detect_per_dior} also show that the `baseball field' and `tennis court' categories reach better detection performance. This is probably because these object two categories have smaller in-category variations. 

\begin{figure*}[hb]
\centering
\includegraphics[width=7in]{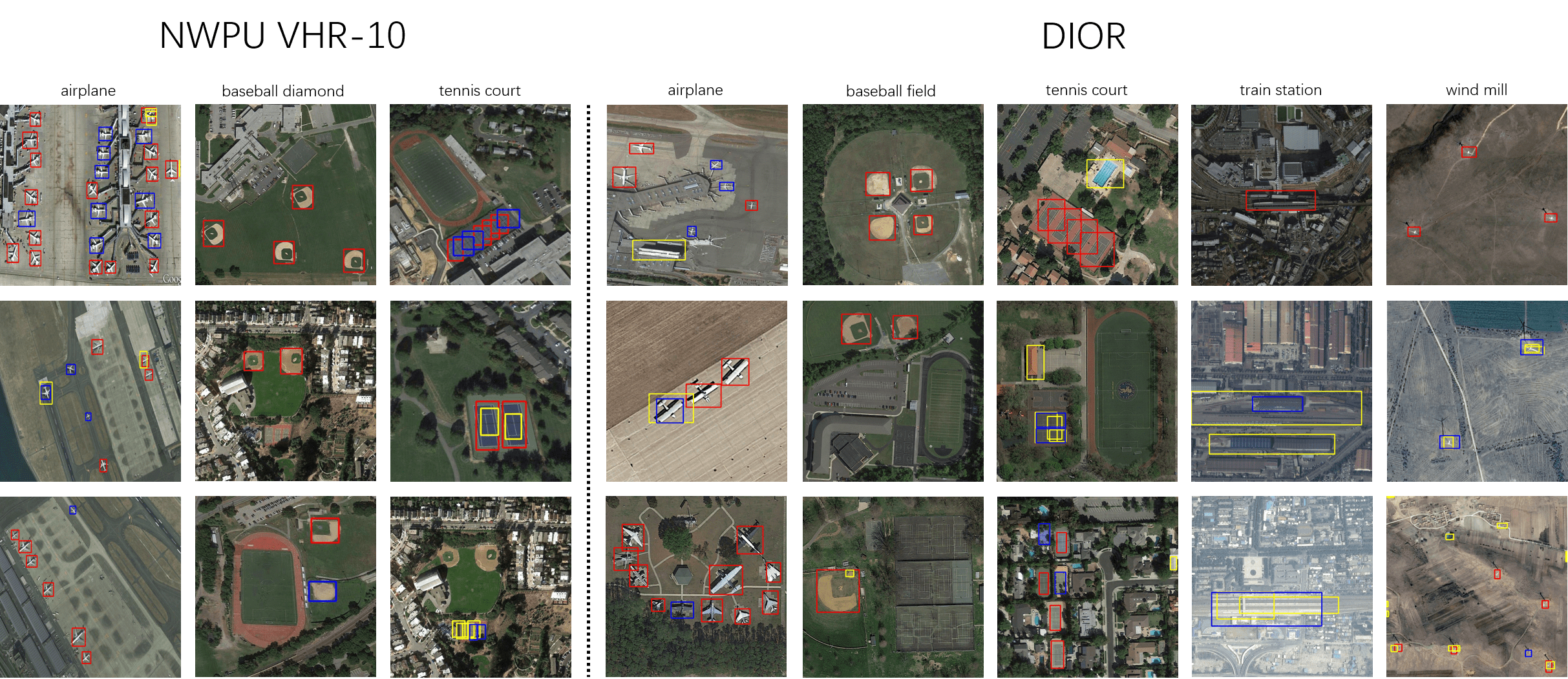}
\caption{Selected examples of our few-shot detection results. Left: detection results on the novel classes of NWPU VHR dataset using 10-shot setting. Right: detection results on the novel classes of DIOR dataset using 20-shot setting. Red, yellow and blue boxes indicate true positive, false positive and false negative detection respectively.}
\label{fig:cases}
\end{figure*}

Fig. \ref{fig:cases} shows some examples of the few-shot detection results of our FSODM model on the NWPU VHR-10 dataset and DIOR dataset. As shown in Fig. \ref{fig:cases}, our model can successfully detect most of the objects in all novel classes of NWPU VHR-10 and DIOR datasets. Most of the failure cases come from the missing or falsely detecting small objects. Moreover, with only a few annotated samples of the novel classes, our model fails to accurately localize `train stations' with large size or appearance variations.

\section{Discussion}\label{sc_discussion}

\subsection{Detection performance on base classes}\label{Base_perform}
A good few-shot object detection model should not only perform well on the novel classes with few annotated samples but also do not sacrifice the performance on base classes. Which means, it should perform as well as the conventional none few-shot-based models when data are abundant. 

Table \ref{tab:detect_per_nwpu} and Table \ref{tab:detect_per_dior} show the performances of our FSODM model and the comparing methods on the base classes of NWPU VHR-10 and DIOR datasets. From Table \ref{tab:detect_per_nwpu} one can see that all three methods get similar performances on the base classes, with slight differences in the mAP values. On the large scale DIOR dataset, our method performs better than another few-shot-based method \cite{kang2019few}. This demonstrates that our proposed method can better maintain the performance on the base classes under the few-shot detection scenario. The performance of our few-shot-based method achieves the same mAP value as the conventional YOLOv3 detection when a large amount of data is given.

\begin{table}[h]
    \centering
    \begin{tabular}{p{30mm}<{\centering} p{10mm}<{\centering} p{10mm}<{\centering} p{10mm}<{\centering}}
    \hline
    Class&FSODM (Ours)&\cite{kang2019few}&YOLOv3\\
    \hline
    ship&0.72&0.77&0.71\\
    storage tank&0.71&0.80&0.68\\
    basketball court&0.72&0.51&0.62\\
    ground track field&0.91&0.94&0.94\\
    harbor&0.87&0.86&0.84\\
    bridge&0.76&0.77&0.80\\
    vehicle&0.76&0.68&0.77\\
    \hline
    mean&0.78&0.76&0.77\\
    \hline
    \end{tabular}
    \caption{Detection performance (mAP) on the base classes of NWPU VHR-10 dataset.}
    \label{tab:detect_per_nwpu}
\end{table}

\begin{table}[h]
    \centering
    \begin{tabular}{p{30mm}<{\centering} p{10mm}<{\centering} p{10mm}<{\centering} p{10mm}<{\centering}}
    \hline
    Class&FSODM (Ours)&\cite{kang2019few}&YOLOv3\\
    \hline
    airport&0.63&0.59&0.59\\
    basketball court&0.80&0.74&0.83\\
    bridge&0.32&0.29&0.28\\
    chimney&0.72&0.70&0.68\\
    dam&0.45&0.52&0.39\\
    expressway service area&0.63&0.63&0.68\\
    expressway toll station&0.60&0.48&0.57\\
    golf course&0.61&0.61&0.63\\
    ground track field&0.61&0.54&0.70\\
    harbor&0.43&0.52&0.43\\
    overpass&0.46&0.49&0.43\\
    ship&0.50&0.33&0.64\\
    stadium&0.45&0.52&0.43\\
    storage tank&0.43&0.26&0.46\\
    vehicle&0.39&0.29&0.41\\
    \hline
    mean&0.54&0.50&0.54\\
    \hline
    \end{tabular}
    \caption{Detection performance (mAP) on the base classes of DIOR dataset.}
    \label{tab:detect_per_dior}
\end{table}

\subsection{Number of shots}\label{fewshot_perform}

\begin{figure}
    \centering
    \includegraphics[width=3.3in]{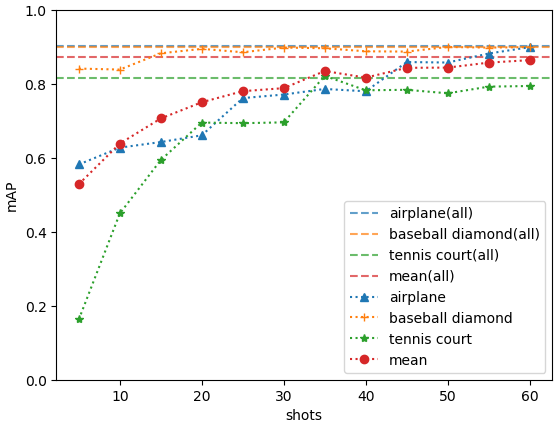}
    \caption{Detection performance on tasks with different shots of NWPU VHR-10 dataset. Horizontal dash lines are performances with all samples in dataset which are 1025 samples for airplane, 519 samples for baseball diamond and 643 samples for tennis court.}
    \label{fig:few_base}
\end{figure}

We investigate the performance of our few-shot object detection model under different numbers of shots (i.e., annotated samples) on the novel classes. To show the advantage of our few-shot based model, we conduct experiments with all training samples from the novel categories of the NWPU VHR-10 dataset and use YOLOv3 as the baseline model. For our FSODM model, we conduct experiments with a larger few-shot range (from 5 shots to 60 shots). As shown in Fig. \ref{fig:few_base}, our model with only 60 (8\%) training samples from the novel categories can achieve almost the same detection performance as the baseline model that uses all training samples. We owe this to the fact that our FSODM model can learn meta-knowledge from base classes and effectively apply it for detection on novel classes, while the baseline model can not well transfer the knowledge from base classes to novel classes. Moreover, on the baseball diamond class, our model with only 20 annotated samples achieves almost the same performance as the baseline model that uses all training samples. This is probably because baseball diamonds have smaller in-category variations and can be easily identified with its structures from a few annotated samples. In contrast, although the airplane class has almost the same detection performance as the baseball diamond class using the baseline model, the few-shot detection performance is significantly worse. This is because objects in the airplane category have larger structural and size variations, as shown in Fig. \ref{fig:cases}, and this challenge impedes our model from getting a satisfying performance with only a few samples (less than 60). Even though, our few-shot-based model can successfully obtain a comparable performance as the baseline model when enough annotated samples (60 shots) are given. 

\subsection{Reweighting vectors}\label{R_coef}

\begin{figure*}
    \centering
    
    \subfigure[]{
        \includegraphics[width=2.2in]{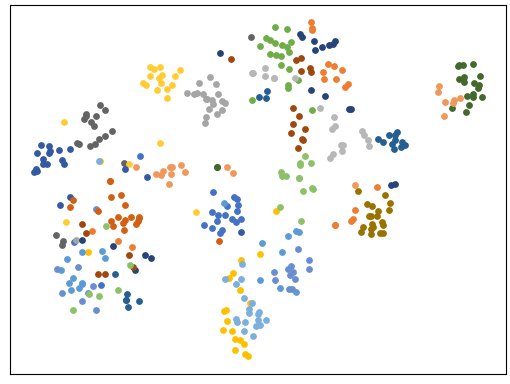}
        \label{fig:reweighting_coef_1}
    }
    \subfigure[]{
        \includegraphics[width=2.2in]{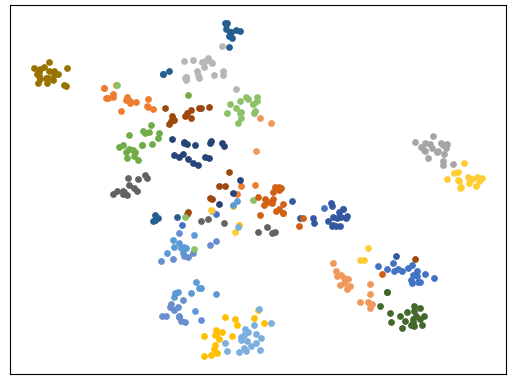}
        \label{fig:reweighting_coef_2}
    }
    \subfigure[]{
        \includegraphics[width=2.2in]{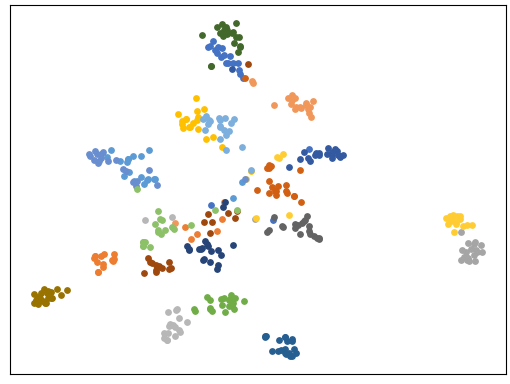}
        \label{fig:reweighting_coef_3}
    }
    \caption{t-SNE \cite{maaten2008visualizing} visualization of reweighting vectors. The reweighting vectors are generated from 400 support images randomly picked from the DIOR dataset (20 images of each category). (a), (b) and (c) are visualizations of reweighting vectors with dimensions of 256, 512 and 1024 respectively.}
    \label{fig:reweighting_coef}
\end{figure*}

In our approach, the reweighting vectors are extracted by the reweighting module and significantly support the final detection performance. To explore the relationship between these reweighting vectors, we use t-Distributed Stochastic Neighbor Embedding (t-SNE) \cite{maaten2008visualizing} to reduce their dimensions and visualize them on the coordinate axis. T-SNE is a dimensionality reduction technique that can pass the inner relationship between high dimension vectors to low dimension vectors. To put it simply, t-SNE keeps close vectors in high dimension space close in low dimension space and remote vectors in high dimension space remote in low dimension space. 

Fig. \ref{fig:reweighting_coef} shows some examples of visualized reweighting vectors. In the figure, reweighting vectors from the same categories tend to aggregate together, which suggests the learned reweighting vectors successfully characterize the object class information from original support masks. In addition, the clustering results in Fig. \ref{fig:reweighting_coef_3} are obviously better than result in Fig. \ref{fig:reweighting_coef_1} and Fig. \ref{fig:reweighting_coef_2}. The reason is that the more elements a reweighting vector has, the more information it carries. Therefore, reweighting vectors with higher dimensions tends to be more capable of representing the object information from support samples.


\section{Conclusions}\label{sc_conclusion}
This paper introduces a new few-shot learning-based method for object detection on remote sensing images, which is among the first to challenge this area. We first formulate the few-shot object detection problem on remote sensing images. Then we introduce our proposed method, which includes three main components: a meta feature extractor, a feature reweighting module, and a bounding box prediction module. Each module is designed in a multi-scale architecture to enable multi-scale object detection. Our method is trained with large-scale data from some base classes and can learn meta-knowledge from base classes and generalize well to novel classes with only a few samples. Experiment on two public benchmark datasets demonstrates the powerful ability of our method for detecting objects from novel classes through a few annotated samples. This work is a very first step in the few-shot detection in remote sensing field and we will further improve it and keep exploring in this field.


\bibliographystyle{IEEEtran}
\bibliography{egbib}

\end{document}